\documentclass{article}
\usepackage{spconf,amsmath,graphicx}
\usepackage{times}
\usepackage{epsfig}
\usepackage{amssymb}
\usepackage{multirow}
\usepackage{multicol}
\usepackage{booktabs}
\usepackage{enumitem}
\usepackage{sectsty}
\usepackage{url}
\sectionfont{\centering}
\usepackage{titlesec}
\titlespacing{\section}{0pt}{2\parskip}{-2\parskip}
\titlespacing{\subsection}{0pt}{2\parskip}{-2\parskip}
\title{Agent-Environment Network for Temporal Action Proposal Generation}
\name{Viet-Khoa Vo-Ho$^{\star\dagger}$\qquad Ngan Le$^{\dagger}$ \qquad Kashu Yamazaki$^{\dagger}$ \qquad Akihiro Sugimoto$^{\mathsection}$ \qquad Minh-Triet Tran$^{\star}$}
\address{$^{\star}$ Faculty of Information Technology, University of Science, VNU-HCM, Vietnam \\
Vietnam National University, Ho Chi Minh City, Vietnam\\
$^{\dagger}$ Department of Computer Science, University of Arkansas, Fayetteville, USA \\
$^{\mathsection}$National Institute of Informatics, Japan}

\begin{document}
%\ninept
%
\maketitle
\begin{abstract}
 Temporal action proposal generation is an essential and challenging task that aims at localizing temporal intervals containing human actions in untrimmed videos. Most of existing approaches are unable to follow the human cognitive process of understanding the video context due to lack of attention mechanism to express the concept of an action or an agent who performs the action or the interaction between the agent and the environment. Based on the action definition that a human, known as an agent, interacts with the environment and performs an action that affects the environment, we propose a \textbf{contextual Agent-Environment Network}. Our proposed contextual AEN involves (i) \textbf{agent pathway}, operating at a local level to tell about which humans/agents are acting and (ii) \textbf{environment pathway} operating at a global level to tell about how the agents interact with the environment. Comprehensive evaluations on 20-action THUMOS-14 and 200-action ActivityNet-1.3 datasets with different backbone networks, i.e C3D and SlowFast, show that our method robustly exhibits outperformance against state-of-the-art methods regardless of the employed backbone network. %We will release the code publicly available upon the acceptance.
\end{abstract}
\vspace{8mm}
\section{Introduction}
% + ADD AN EXAMPLE WHERE WE CAN SEE THAT AGENT AND ENVIRONMENT EFFECT TO THE ACTION PROPOSAL GENERATION
%  + NONE OF THE PREVIOUS WORK HAS EXPLOITED AND UTILIZED THE RELATIONSHIP BETWEEN AGENT WHO PERFORMS ACTION AND ENVIRONMENT WHICH RECEIVE THE ACTION
 
 Temporal action proposal generation (TAPG) aims at proposing video temporal intervals that likely contain an action in an untrimmed video with both action categories and temporal boundaries. This task has promising applications, such as action recognition \cite{SlowFast}, summarization \cite{summarization_2015, summarization_2016}, captioning \cite{captioning_2019, captioning_2020}, and video recommendation \cite{video_recommendation_1}. A robust TAPG method should be able to (i) generate temporal proposals with boundaries covering action instances precisely and exhaustively, (ii) cover multi-duration actions, and (iii) generate reliable confidence scores to retrieve proposals properly. Despite many recent endeavors, TAPG remains an open problem, especially when facing real-world complications such as action duration variability, activity complexity, camera motion, and viewpoint changes.

The limitations of the existing TAPG can be summarized as follows:
\begin{itemize}[leftmargin=*, wide=0pt, topsep=0pt]
\setlength{\itemsep}{0pt}
\setlength{\parskip}{0pt}
\item Most of existing work \cite{C3D}, \cite{2_stream_1, 2_stream_2}, \cite{SlowFast} extracts video visual representation by applying a backbone model into whole spatial dimensions of video frames. This tends predictions over-biased towards the environment rather than agents committing actions because the agents together with their actions usually occupy a small region compared to the entire frame.
\item Existing approaches treat everything in a video frame in the same manner and does not pay attention to the difference among three key entities, i.e., agent, action, and environment, for temporal action proposal. Attention mechanism that enables us to capture such different key entities as well as to express the relationship between them is missing.
\item Most of the existing approaches are unable to follow the human cognitive process of understanding the video content. In the human cognitive process, a person focuses on deciding what an agent is doing through the observation of agent activities and the environment around the agent. Nevertheless, such a process is not taken into account at all. Instead, existing work just applies a backbone network into entire spatial dimensions of video snippets of frames (8-frame snippets or 16-frame snippets, etc.).
\end{itemize}
%(i) Although lately new boundary based methods  \cite{lin2018bsn, liu2019multi} achieve state-of-the-art performances with a large margin from anchor based methods, they still use a rather abstract (poor) \textcolor{red}{replace another word} input features, specifically the output from softmax layer of Two-Stream model \cite{2_stream_1}, which is trained on the ActivityNet dataset \cite{caba2015activitynet} for recognizing one of two hundred actions \textcolor{red}{need citation. Need to rewrite}. This leads to poor performance on complicated actions like ... \textcolor{red}{need examples and citation}. Besides, such networks extract visual representation by applying backbone model, i.e. 3D CNN network \cite{C3D} to the whole spatial dimensions of the video, which makes the predictions being biased to the environment instead of agents committing actions. Working in this fashion may negatively affect the results if the actions appear in small regions or are not relating to the environment. (ii) The utilization of 3D CNN networks as a backbone for capturing features of videos to later use on vary tasks are very common, especially in temporal action proposal. However, they are superficially used by applying the network on the whole spatial dimensions of video snippets of frames (8-frame snippets or 16-frame snippets etc.). This may lead to the failure of proposing actions in situations where the environment semantically differs from the action or the actions are committed in small regions.

To address the above drawbacks, we propose a novel \textbf{contextual AEN} to semantically extract video representation. Our proposed AEN contains two semantic pathways corresponding to (i) \textbf{agent pathway} and (ii) \textbf{environment pathway}. The contribution of
\textbf{contextual AEN} is two-fold:

\begin{figure*}[ht!]
\centering
  \includegraphics[width=0.8\linewidth]{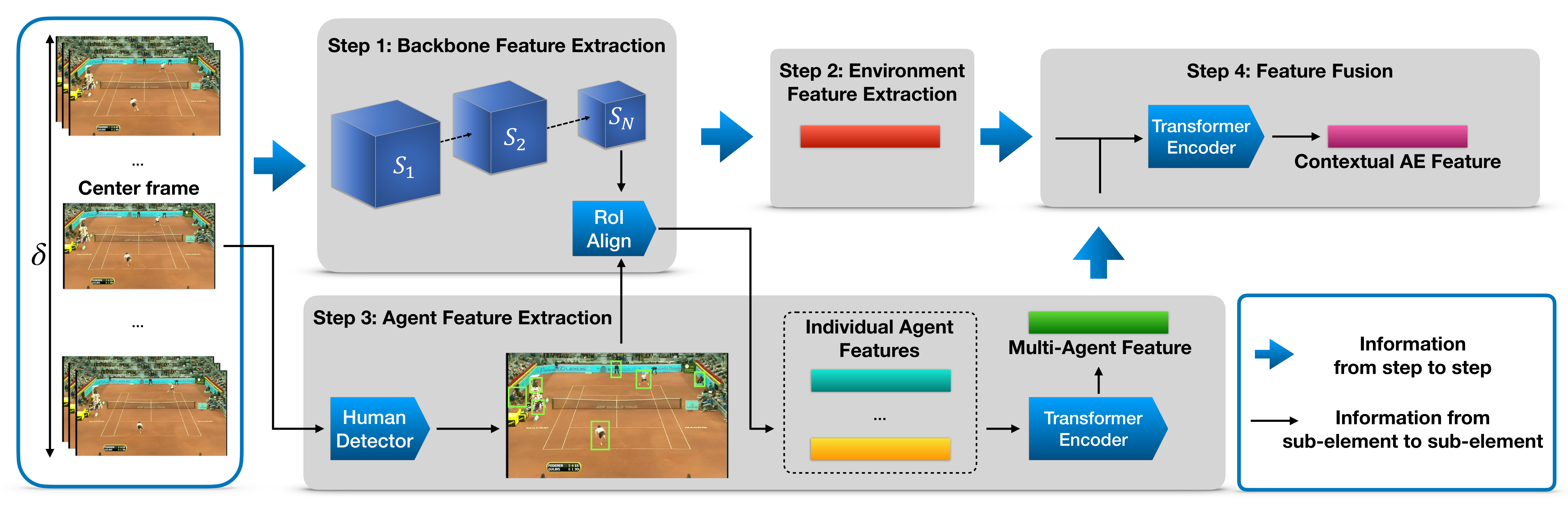}
  \caption{The architecture of our proposed contextual Agent-Environment (AE) representation network (AERN).}
  \label{video_representation}
\end{figure*}

\begin{itemize}[leftmargin=*, wide=0pt, topsep=0pt]
\setlength{\itemsep}{0pt}
\setlength{\parskip}{0pt}
    \item AEN includes (i) Agent-Environment representation network (AERN) to extract rich features sequence from an untrimmed video and (ii) boundary matching networks to evaluate confidence scores of densely distributed proposals generated from the extracted feature.
    \item A novel video contextual Agent-Environment (AE) visual representation is introduced. Our semantic AE visual representation involves two parallel pathways to represent every snippet of the video: (i) agent pathway, operating at a local level to tell what the agents in the snippet are doing and which agents deserve to be concentrated more on; (ii) environment pathway, operating at a global level to express the relationship between the agents and the environment. These two pathways are fused together by our attention mechanism for the video representation where a feature may focus more on either local or global levels entirely depending on the context of its corresponding snippet.
%    \item We benchmark the proposed AEN on two popular datasets, i.e. ActivityNet-1.3 and THUMOS-14, with two different backbone networks i.e. C3D and SlowFast. Our proposed AEN  has achieved state-of-the-art performance on both datasets regardless of backbone networks used.
\end{itemize}

\section{Related Work}

TAPG \cite{2_stream_1, SlowFast, actionproposal_2016, FasterR_CNN_Action, anchor_1, anchor_2, anchor_3, boundary_0, lin2018bsn, liu2019multi} aims at proposing intervals so that each of them contains an action instance with its associated temporal boundaries and confidence score in untrimmed videos. There are two main approaches in TAPG: anchor-based and boundary-based. The anchor-based methods \cite{actionproposal_2016, FasterR_CNN_Action, anchor_1, anchor_2, anchor_3} are inspired by anchor-based object detectors in still images like Faster R-CNN \cite{FasterRCNN}, RetinaNet \cite{RetinaNet}, or YOLO \cite{yolov3}. These methods deal with the proposal task as a classification task 
where multiple predefined anchors with different lengths are regarded as classes and a class that best fits the ground truth action length is used as ground truth true class for training. 
Although this approach helps to save computational costs, it lacks the flexibility of action duration. 
The boundary-based methods \cite{boundary_0, lin2018bsn, liu2019multi}, on the other hand, break every action intervals into starting and ending points and learn to predict them. In the inference phase, starting and ending probabilities at every timestamp in the given video are predicted. Then, points with local peak in probability are chosen as potential boundaries. The potential starting points are paired with potential ending points for a potential action interval when their interval fits in the predefined upper and lower threshold, along with a confidence score being a multiplication of the starting and ending probabilities.  As one of the first boundary-based methods, \cite{boundary_0} defined actionness scores by grouping continuous high-score regions as a proposal. Later, \cite{lin2018bsn} proposed a two-stage strategy where boundaries and actionness scores at every temporal point are predicted in the first stage and fused together, filtered by Soft-NMS to get the final proposals at the second stage. 
\cite{liu2019multi} improved \cite{lin2018bsn} by generating a boundary-matching matrix instead of actionness scores to capture an action-duration score for more descriptive final scores.

\section{Proposed Method}

% Given an untrimmed video $\mathcal{V} = \{x_l\}^{L}_{l=1}$ with $L$ frames, our goal is to generate a set of temporal segments which possibly and tightly contain actions. Given a set of ground truth action segments $\mathcal{A} = \{a_i=(s_i, e_i)\}^{M}_{i=1}$ having $M$ temporal segments with an action segment comprised of a starting timestamp ($s_i$) and an ending timestamp ($e_i$), our objective is formalized by minimizing the following objective function:

%\begin{equation}
%\begin{split}
%& \mathcal{L} =\sum\limits_{i=1}^M -\log p(a_i|\mathcal{V})\\
%& =\sum\limits_{i=1}^M - \log p(s_i|\mathcal{V}) - \log p(e_i|\mathcal{V}) - \log p(e_i - s_i|s_i)
%\end{split}
%\label{eq:objective}
%\end{equation}

% As proposed by \cite{lin2018bsn} and \cite{actionproposal_2019}, the above objective function may also be achieved in an indirect way. Concretely, the above objective function can be formulated as in Eq.\ref{eq:objective2} by decomposing the action proposal generation problem into detecting starting, ending timestamps of every actions together with their duration:

% \begin{equation}
% \mathcal{L}=\sum\limits_{i=1}^M - \log p(s_i|\mathcal{V}) - \log p(e_i|\mathcal{V}) - \log p(e_i - s_i|s_i)
% \label{eq:objective2}
% \end{equation}

%\subsubsection{Formulation for Video Representation}
Given an untrimmed video $\mathcal{V} = \{x_l\}^{L}_{l=1}$ with $L$ frames, our goal is to generate a set of temporal segments, each of which possibly and tightly contain an action. Let us denote $F$ as the visual representation of video $\mathcal{V}$, which is firstly divided into $T=\left \lfloor{\frac{L}{\delta}}\right\rfloor$ non-overlapping $\delta$-frame snippets. Let $\phi$ be a feature extraction function which is applied to each $\delta$-frame snippet, the visual representation $F$ is then defined as follows:
\begin{equation}
\begin{split}
F =\{f_i\}^{T}_{i=1}  =\{\phi(x_{\delta\cdot(i-1)+1},...,x_{\delta\cdot i})\}^{T}_{i=1}
\label{eq:funF}
\end{split}
\end{equation}

In the next two subsections, we discuss how we devise Agent-Environment Representation Network (AERN) as a function $\phi$ and how we integrate it with an action proposal generation module, respectively.

%By applying a function $\phi$ as a represent function to the video $\delta$-frame snippets, we have the below feature list:

%$$F=\{f_i=\phi(x_{\delta*(i-1)+1},...,x_{\delta*i})\}^{\left \lfloor{\frac{L}{\delta}}\right \rfloor}_{i=1}$$

% In most of the previous works \cite{TCN, MSRA, Prop-SSAD, CTAP, SRG, lin2018bsn, liu2019multi, actionproposal_2019}, $\phi$ is simply defined as a feature vector from a hidden layer of C3D Network \cite{C3D}, two-stream Network \cite{2_stream_1}, or Slow-Fast \cite{SlowFast} given a $\delta$-frame snippet. However, as stated in the above sections, this may lead to capture insufficient information or noise information because actions and the agents who create the actions usually take place in small spatial regions of the video. Hence, in this work we propose a novel video contextual visual representation, named Agent-Environment network (AEN), which incorporates both agents, their actions and the interaction between agents and their environment. Our proposed AEN with new feature extraction mechanism can be developed by any backbone such as C3D Network \cite{C3D}, Two-Stream Network \cite{2_stream_1}, or the latest model of SlowFast Network \cite{SlowFast}. %More details are discussed in section \ref{sub:CAE feature}.

\subsection{AE Representation Network(AERN)}
\label{sub:CAE feature}
%Our proposed AEN consists of two components and is demonstrated in Fig. \ref{proposed_network}. The first component, \textbf{Contextual AE representation network}, extracts contextual AE visual representation of a $\delta$-frame snippet at both global and local levels.
%and is detailed in section \ref{subsec:representation}. 
%The second component, \textbf{boundary matching network}, takes the first component as the input and generates the action proposals.
%and is briefly described in section \ref{subsec:BMN}.

%\subsubsection{Contextual AE Representation Network}
%\label{subsec:representation}
Our proposed AERN extracts contextual AE visual representation of a $\delta$-frame snippet at both global and local levels, which plays a key role in temporal action proposals generation. Considering our goal is extracting features for a $\delta$-frame snippet from frame $t$ to frame $t + \delta$, the AERN is illustrated in Fig.\ref{video_representation}(a) and consists of following steps:
%Notably, there are two kinds of feature extracted in our proposed network, i.e., environment feature plays a role of global semantic level and is extracted through step 2 and agent feature plays a role of local semantic level and is extracted through step 3.

\noindent 
\textbf{Step 1: Backbone Feature Extraction:}
In action recognition, a 3D convolutional backbone network is usually used to encode global semantic information of a $\delta$-frame snippet. 
%The snippet of frames may go through several blocks of convolutional layers to extract spatiotemporal features before being pooled into a single vector and passed through several fully connected layers for classification.
In this work, we employed C3D \cite{C3D} and SlowFast \cite{SlowFast} pre-trained on Kinetics-400 \cite{Kinetics} as our backbone feature extractor. In order to capture enough semantic information of the snippet while keeping enough resolution in spatial domains, we discard the last fully connected layers to use the feature map $S_N$ from the last convolutional block, which is crucial in Step 3.

\noindent 
\textbf{Step 2: Environment Feature Extraction:}
To extract the environment feature, feature map $S_N$ is passed through average pooling and several fully connected layers until the softmax layer, outputting a vector containing semantic information of the overall scene, namely, environment feature $\phi_e$. This pathway captures the information at the global level of the scene, however, it may not capture small details like the motions of humans.

%Because all spatial dimensions are processed, this pathway captures the abstract information at global level of the scene, however, it may not capture details like motions of humans.

\noindent 
\textbf{Step 3: Multi-Agent Feature Extraction:}
An agent in a $\delta$-frame snippet is denoted as a human appears in it. To detect all agents existing in a $\delta$-frame snippet, we start with the center frame by applying a human detector.
%Firstly, we search for these agents by feeding the center frame of $\delta$-frame snippet into a human detector to detect bounding boxes containing humans that appeared in this frame and hopefully still appear in the surrounding frames. 
These bounding boxes around agents are then used to guide the RoIAlign \cite{MaskRCNN_ICCV17} to extract local features from $S_N$. Each local feature corresponds to an agent and all local features from multiple agents are fused into a single multi-agent feature $\phi_a$ via an attention model based on Transformer Encoder \cite{attention_is_all_you_need}. Thus, we obtain an agent $\phi_a$ from multiple agents of $\delta$-frame snippet. 
%, forming a list of agents features, with each feature corresponds to an individual human presenting in the video frame.
%Then, every the agent feature in the features list is fused together into a single multi-agent feature $\phi_a$ via an attention model. The attention model looks at all agents and assigns up-weights to agents who play important roles in the video or are committing observed actions and vice versa. The agents features are then weighted summed to form $\phi_a$.

In this step, we adopt a Faster R-CNN \cite{FasterRCNN} pre-trained on MS-COCO \cite{cocodataset} as our human detector because of its good performance and popularity.
%The human detector works as a hard attention module that eliminates background regions and only emphasize on humans or agents moving in the scene. On the other hand, the attention model works as a soft attention module that helps to concentrate on the right agents but also keeps information of the other agents because the activities we observe may requires the interaction between these agents.

\noindent 
\textbf{Step 4: Feature Fusion:}
In our proposed AEN, environment feature $\phi_e$ plays at the global level while multi-agent feature $\phi_a$ plays at the local level. 
%Finally, the environment feature $\phi_e$ and multi-agent feature $\phi_a$ are fused by another attention model. 
After simultaneously extracting these features, Transformer Encoder \cite{attention_is_all_you_need} is employed to re-weight them by a proper ratio, which helps the overall model to know which information to consider while reasoning the action proposals, i.e. deciding whether to emphasize on detailed information of agents or overall information of the scene.

\begin{figure}[!t]
  \centering
  \includegraphics[width=0.88\linewidth]{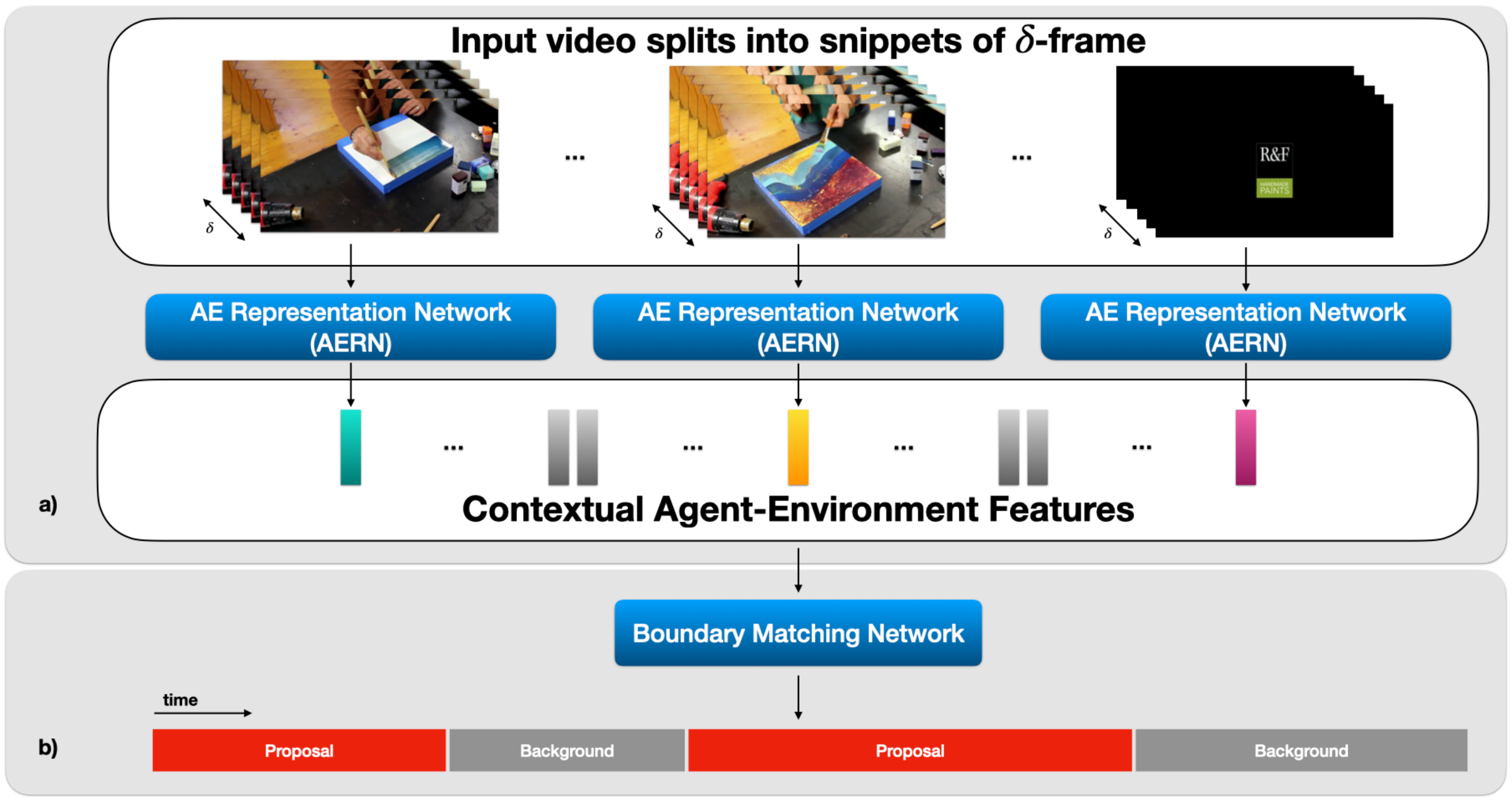}
  \caption{An overview architecture of our proposed AEN for action proposal generation where AE Representation Network is in Fig.\ref{video_representation} and described in section \ref{sub:CAE feature}}
  \label{proposed_network}
\end{figure}

\subsection{Deployment with Action Proposal Network}
\label{Overall}

Our proposed AEN is easily deployed and incorporated with any TAPG network in an end-to-end framework as shown in Fig. \ref{proposed_network}. In this paper, Boundary-Matching Network (BMN) is employed because of its impressive performance. BMN is a fully convolutional network with 3 modules, namely, Base Module (BM), Temporal Evaluation Module (TEM), and Proposal Evaluation Module (PEM). BM processes the input features through several 1D convolutional layers, producing output features that are fed into TEM and PEM simultaneously. TEM aims to produce the probabilities for every temporal point in the features set being a starting or ending boundaries. Meanwhile, PEM produces two matrices, each of which densely contains the confidence scores of every possible duration at every starting temporal point, but are trained by two different types of loss functions as suggested by \cite{actionproposal_2019}. 
%These matrices would have a shape of $D \times T$ with $D$ is the maximum length of the proposals in snippets that we consider and $T$ is the number of snippets. In this work, we set $D = T$ as suggested by \cite{actionproposal_2019}.
\begin{table*}[ht]
\centering
\caption{Comparison in terms of AR@AN and AUC on validation set and test set of ActivityNet-1.3 dataset}
\footnotesize
\begin{tabular}{|c|c|c|c|c|c|c|c|c|c|c|c|}
\hline 
\multicolumn{1}{|c|}{\multirow{2}{*}{}} & \multicolumn{1}{c|}{\multirow{2}{*}{\begin{tabular}[c]{@{}c@{}}TCN\\  \cite{TCN}\end{tabular}}} & \multicolumn{1}{c|}{\multirow{2}{*}{\begin{tabular}[c]{@{}c@{}}MSRA\\ \cite{MSRA}\end{tabular}}} & \multicolumn{1}{c|}{\multirow{2}{*}{\begin{tabular}[c]{@{}c@{}} Prop-SSAD\\ \cite{Prop-SSAD}\end{tabular}}} & \multirow{2}{*}{\begin{tabular}[c]{@{}l@{}}CTAP\\ \cite{CTAP}\end{tabular}} & \multirow{2}{*}{\begin{tabular}[c]{@{}l@{}} BSN\\  \cite{lin2018bsn}\end{tabular}} &
\multirow{2}{*}{\begin{tabular}[c]{@{}l@{}} SRG\\  \cite{SRG} \end{tabular}} &
\multirow{2}{*}{\begin{tabular}[c]{@{}l@{}}MGG\\ \cite{liu2019multi} \end{tabular}} &
\multirow{2}{*}{\begin{tabular}[c]{@{}l@{}}BMN\\  \cite{actionproposal_2019} \end{tabular}} &
\multicolumn{2}{c|}{Our AEN} \\ \cline{10-11}
\multicolumn{1}{|c|}{}  & \multicolumn{1}{c|}{}  & \multicolumn{1}{c|}{} &
\multicolumn{1}{c|}{} & & & & & &    SlowFast   &   C3D  \\ \hline

            %  & TCN \cite{TCN}  & MSRA\cite{MSRA}  & Prop-SSAD \cite{Prop-SSAD} & CTAP \cite{CTAP}  & BSN \cite{lin2018bsn}   & MGG \cite{liu2019multi}  & BMN \cite{actionproposal_2019}   & \textbf{AEN (Ours)}           \\ \hline
AR@100 (val) &  -     &   -    & 73.01     & 73.17 & 74.16 & 74.65 & 74.54 & 75.01 & 75.62   &      \textbf{75.65}        \\ \hline
AUC (val)          & 59.58 & 63.12 & 64.40     & 65.72 & 66.17 & 66.06 & 66.43 & 67.10 & 67.78  &  \textbf{68.15} \\ \hline
AUC (test)         & 61.56 & 64.18 & 64.80     &  -     & 66.26 & - & 66.47 & 67.19 & 68.45       &    \textbf{68.99}     \\ \hline
\end{tabular}
\label{activitynet}
\vspace{-5mm}
\end{table*}

\begin{table}[b]
\centering
\vspace{-5mm}
\caption{Generalization evaluation on ActivityNet 1.3.}
\footnotesize
\begin{tabular}{l c c c c}
\toprule
 & \multicolumn{2}{c}{Seen} & \multicolumn{2}{c}{Unseen} \\
\hline
Training Data & AR@100 & AUC & AR@100 & AUC \\
\hline

Seen+Unseen & 74.58 & 66.96 & 75.25 & 67.49 \\
%Seen        & 74.07 & 66.47 & \textbf{74.14} & \textbf{66.36} \\
Seen        & 74.40 & 66.69 & \textbf{73.66} & \textbf{65.92} \\

\toprule
\end{tabular}
\label{generalization}
\end{table}

\subsection{Training Phase}
\noindent 
\textbf{Label Generation:}
We follow \cite{actionproposal_2019, lin2018bsn} to generate the ground truth labels for the training process including starting labels, ending labels for TEM training and duration labels for PEM training. The starting and ending labels are generated for every snippet of the video, which are called $L_S=\{l^s_n\}_{n=1}^T$ and $L_E=\{l^e_n\}_{n=1}^T$, respectively. A label point $l^s_n$ (or $l^e_n$) is set to $1$ if its corresponding timestamp in the video is the nearest to any ground truth starting (ending) timestamp.

% \begin{equation}
%     l^{\text{gt}}_{n} = 
% \begin{cases}
%     1,& \text{if } \sum\limits^{R_{\text{gt}}}_{r_{\text{gt}}} \frac{r_{n} \cap r_{\text{gt}}}{r_{\text{gt}}} \geq 0.5\\
%     0,              & \text{otherwise}
% \end{cases}
% \end{equation}

%where $R_{\text{gt}}$ can be replaced by $R_S$ or $R_E$, and $\cap$ operator returns the intersection between two regions.
The duration labels for a video are gathered into a matrix $L_D \in [0, 1]^{D \times T}$ where $D$ is the maximum length of proposals being considered in number of snippets, as suggested in \cite{actionproposal_2019}, we set $D=T$ and $D=T/2$ for experiments on ActivityNet-1.3 \cite{caba2015activitynet} and THUMOS-14 \cite{THUMOS14}, respectively. With an element at position $(t_i, t_j)$ stands for a proposal action $a_p=(t_s=\frac{t_j\cdot T}{t_v}, t_e=\frac{(t_j+t_i)\cdot T}{t_v})$, it is assigned by $1$ if its Interaction-over-Union with any ground truth action in $\mathcal{A}=\{a_i\}_{i=1}^{M}$ reach a local maximum, or $0$ otherwise.

\noindent
\textbf{Loss function:}
As mentioned in section \ref{Overall}, TEM generates probabilities vectors of starting and ending boundaries ($P_S$ and $P_E$), while PEM generates two actionness scores matrices $P^{cc}_D$ and $P^{cr}_D$. These four outputs are trained simultaneously by different loss functions as following:
\begin{equation}
    \mathcal{L}_{TEM} = \mathcal{L}_{binary}(P_S, L_S) + \mathcal{L}_{binary}(P_E, L_E)
\end{equation}
\begin{equation}
    \mathcal{L}_{PEM} = \mathcal{L}_{binary}(P^{cc}_D, L_D) + \lambda_{reg} \cdot \mathcal{L}_{2}(P^{cr}_D, L_D)
\end{equation}
\begin{equation}
    \mathcal{L} = \lambda_1 \cdot \mathcal{L}_{TEM} + \lambda_2 \cdot \mathcal{L}_{PEM}
\end{equation}

We follow \cite{actionproposal_2019, lin2018bsn} and set $\lambda_{reg} = 10$ and $\lambda_1 = \lambda_2 = 1$. Furthermore, $\mathcal{L}_{binary}$ is a weighted binary log-likelihood function to deal with imbalanced number of negative and positive examples in groundtruth labels:

\vspace{-0.5cm}
\begin{equation}
    \mathcal{L}_{binary} = \frac{1}{N} \sum\limits^{N}_{i=1} \alpha^{+} \cdot l_i \cdot \log{p_i} + \alpha^{-} \cdot (1 - l_i) \cdot \log{p_i},
\end{equation}
where $l_i$ and $p_i$ are label and probability of the output, respectively. $\alpha^{+} = \frac{N}{N^{+}}$ and $\alpha^{-} = \frac{N}{N^{-}}$, with $N$, $N^-$ and $N^+$ are total number of examples and total number of positive and negative examples, respectively.

\subsection{Inference Phase:}
During inference, four outputs are generated by BMN model \cite{actionproposal_2019} from features set extracted by our AEN, including $P_S$, $P_E$, $P^{cc}_D$, and $P^{cr}_D$. Peaking probabilities of starting and ending boundaries from $P_S$ and $P_E$, which are local maximums, are selected to form initial proposals by pairing every peak starting point with peak ending points behind them and within a pre-defined range. For a proposal formed by $t_s$ and $t_e$ boundaries with duration $d_p = t_e - t_s$, its score $score_p$ are computed by the following formula as proposed in \cite{actionproposal_2019}:
\begin{equation}
    score_{p} = P_S[t_s] \cdot P_E[t_e] \cdot \sqrt{P^{cc}_D[d_p, t_s] \cdot P^{cr}_D[d_p, t_s]}
\end{equation}

Then, with a list of proposals and their scores, we apply a Soft-NMS \cite{SoftNMS} to eliminate highly overlapped proposals before outputting the final list of proposals.

\begin{table}[t]
\centering
\caption{Comparison on THUMOS-14 test set (AR@AN).}
\footnotesize
\begin{tabular}{c c c c c c }
\toprule
 Methods & @50 & @100 & @200 & @500 & @1000              \\
\hline
\hline

SCNN-prop \cite{anchor_2} & 17.22 & 26.17 & 37.01 & 51.57 & 58.20        \\
SST \cite{SST_CVPR2017} & 19.90 & 28.36 & 37.90 & 51.58 & 60.27        \\
TURN \cite{anchor_3} & 19.63 & 27.96 & 38.34 & 53.52 & 60.75        \\
MGG \cite{liu2019multi} & 29.11 & 36.31 & 44.32 & 54.95 & 60.98        \\
BSN \cite{lin2018bsn} & 29.58 & 37.38 & 45.55 & 54.67 & 59.48        \\
BMN \cite{actionproposal_2019} & 32.73 & 40.68 & 47.86 & 56.42 & 60.44     \\
%& DBG+SNMS  &    30.55 & 38.82 & 46.59 & 56.42 & 62.17     \\ 
%DBG  &    32.55 & 41.07 & 48.83 & 57.58 & 59.55     \\ \cline{1-6}
Our AEN &\textbf{33.36}&\textbf{42.93}&\textbf{50.34}&\textbf{59.10}&\textbf{64.03}\\
%\hline
%\hline
%\multirow{7}{*}{2Stream}
%& TAG       & 18.55 & 29.00 & 39.61 & - & -                \\
%& CTAP      & 32.49 & 42.61 & 51.97 & - & -                \\
%& MGG       & 39.93 & 47.75 & 54.65 & 61.36 & 64.06        \\
%
%& BSN & 37.15 & 46.75 & 54.84 & 62.19 & 65.22          \\
%& BMN &    37.15 & 46.75 & 54.84 & 62.19 & 65.22     \\
%& DBG+SNMS &    37.32 & 46.67 & 54.50 & 62.21 & 66.40     \\
%& DBG &    40.89 & 49.24 & 55.76 & 61.43 & 61.95     \\ \cline{2-7}
%& Our AEN &     &      &      &      &       \\
\toprule
\end{tabular}
\label{thumos}
\vspace{-3mm}
\end{table}
\section{Experiments}
\subsection{Datasets}
% We evaluate our proposed method on two benchmark datasets, namely ActivityNet-1.3 \cite{caba2015activitynet} and THUMOS-14 \cite{THUMOS14}. 

\noindent
\textbf{ActivityNet-1.3} \cite{caba2015activitynet} is a large scale dataset for human activity understanding, containing roughly 20K untrimmed videos which are divided into training, validation and test sets with the ratio of 0.5, 0.25 and 0.25, respectively.

%Every video is annotated with one or more temporal intervals accommodating any activity out of 200 activities of interest.
%We re-scaled all videos to 1600 frames by linear interpolation and extracted features for every separate snippet with length $\delta = 16$ frames. Therefore, every video sequence will be represented by a feature sequence with the length of exactly 100 features.

%Due to the unavailability of annotations on test split, we compare and report performances of our approach and other state-of-the-art methods on the validation set, unless otherwise stated.
\noindent 
\textbf{THUMOS-14} \cite{THUMOS14} is primarily a dataset for action recognition, yet, it also opens the action localization track, which is held on a portion of its validation set for training and another portion of test set for testing, with each comprised of 200 and 214 videos, respectively; and captures 20 different actions.

For comparability purposes, we follow the same settings as it was in \cite{actionproposal_2019} for both datasets.

%Following \cite{SST_CVPR2017, SSTAD_BMVC17}, we randomly split the training set into training and validation sets with the ratio of 4:1 to do hold-out evaluation while training.

%Due to the lengthy attribution of videos in THUMOS-14 \cite{THUMOS14}, we used a fixed size observation window to stride through every video and collect it whenever it overlaps a ground truth action interval. By doing this, the processed dataset will be more balanced between the background and action duration. In this paper, we set the window size to $l_w = 2048$ and stride step to $s_w = 1024$, with the snippet size $\delta = 16$ we will have video representation as a feature sequence of size 128, which equals to the experimental setup of \cite{actionproposal_2019}.

\subsection{Implementation Details}
For ActivityNet-1.3, we benchmark our proposed AEN with both C3D \cite{C3D} and SlowFast \cite{SlowFast} backbone, whereas, for THUMOS-14, we only benchmark our method on C3D backbone. All backbones are pre-trained on Kinetics-400 \cite{Kinetics}. Following \cite{actionproposal_2019, lin2018bsn}, we trained our proposed network with Adam update rule is employed with the initial learning rate of 0.0001 and 0.001 for ActivityNet-1.3 and THUMOS-14, respectively.
%The backbone feature map $S_4$ of each backbone are 2048 dimensions and 2304 dimensions, respectively. 
%We always keep this size of the feature through out our proposed network and output the Contextual Agent-Environment Feature of size the same as backbone feature size.

%The Transformer Encoders we used for contextually merging agent features together or merging multi-agent feature with environment feature together share the same architecture of 4 attention heads and 1 transformer layer.

%We trained our proposed network for 10 epochs and 20 epochs on ActivityNet-1.3 and THUMOS-14, respectively. Besides, Adam update rule is employed with the initial learning rate of 0.0001 and 0.001 for ActivityNet-1.3 and THUMOS-14, respectively.

\subsection{Experimental Results}
%In evaluating the proposals, we follow the evaluation paradigm for the temporal localization task in ActivityNet-1.3 dataset. We adopt different IoU thresholds to calculate the average recall (AR) with average number of proposals (AN). We use the Area Under the AR vs AN curve (AUC) at 100 proposals per video which is averaged across ten different IoU thresholds uniformly. A set of IoU thresholds [0.5:0.05:0.95] is used on ActivityNet-1.3, while a set of IoU thresholds [0.5:0.05:1.0] is used on THUMOS-14. For ActivityNet-1.3, area under the AR vs AN curve (AUC) is also used as the evaluation metrics.

Table \ref{activitynet} shows the comparison in terms of AR@AN (AN = 100) and AUC between our AEN against other SOTA methods on both validation and test sets of ActivityNet-1.3 dataset. Compared to SOTA approaches, our AEN obtains better performance with large margins on both AR@AN and AUC metrics regardless of the backbone networks. Likewise, our AEN also gives a superior performance on THUMOS-14 in Table \ref{thumos} when compared to SOTA approaches on this dataset.

Generalization is also an important aspect to be evaluated in TAPG. We conduct experiments on ActivityNet-1.3 \cite{caba2015activitynet} to evaluate this property, in which videos in two non-overlapped action class subsets of "Sports, Excercise, and Recreation" and "Socializing, Relaxing, and Leisure" are collected into \textit{Seen} and \textit{Unseen} subsets, respectively. Table \ref{generalization} delivers two training settings, results evaluated on Unseen subset does not drop significantly when training only on \textit{Seen} subset comparing to training on \textit{Seen+Unseen} sets, which implies that our AEN achieves high generalizability in generating proposals.

\begin{figure}[t]
\centering
  \includegraphics[width=0.95\linewidth]{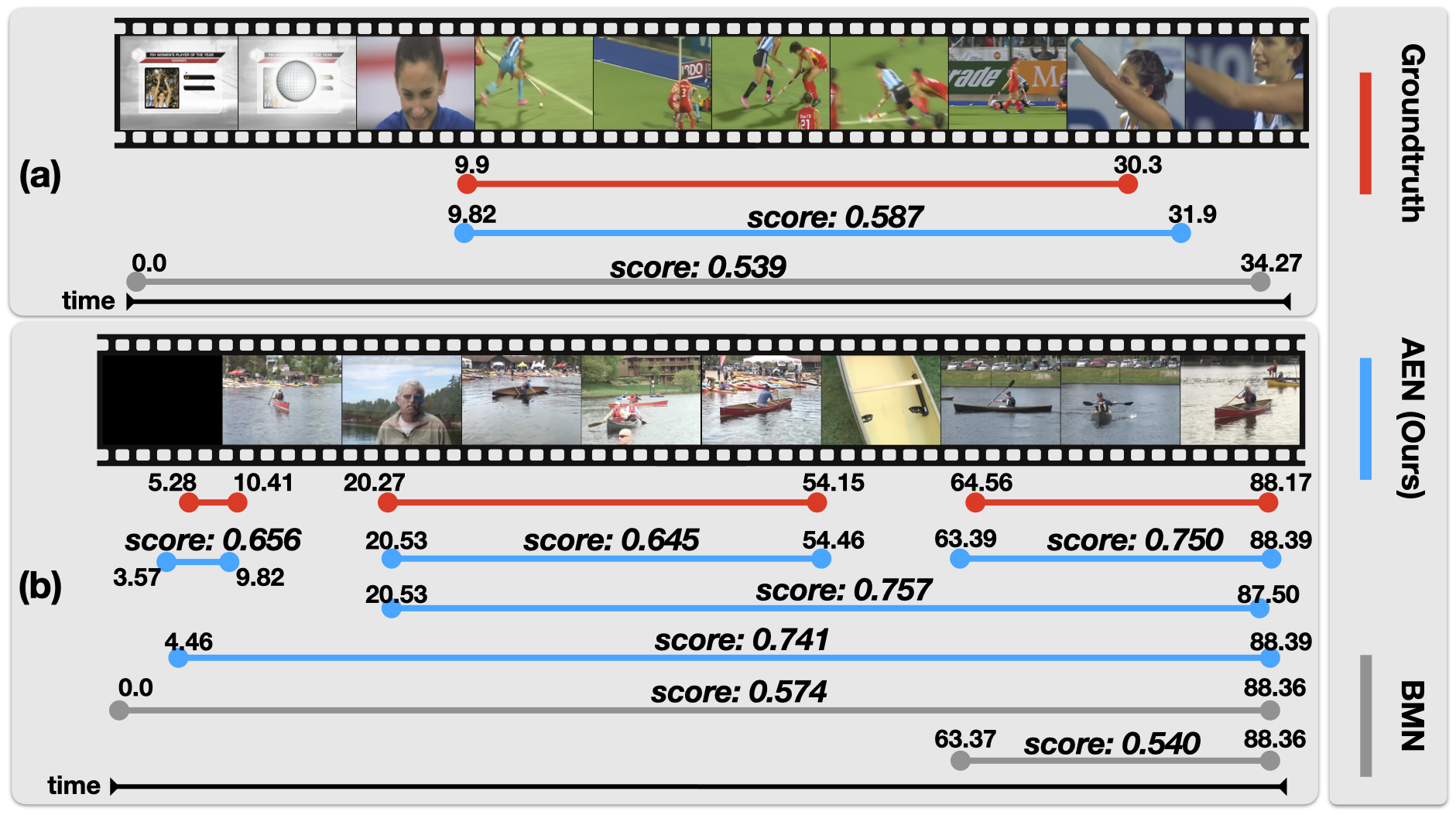}
  \vspace{-3mm}
  \caption{Qualitative results by BMN \cite{actionproposal_2019} and our proposed AEN on ActivityNet-1.3 \cite{caba2015activitynet} on C3D backbone network.}
  \label{illustration}
\vspace{-5mm}
\end{figure}

\section*{Conclusion}
This paper proposed a novel AEN for the TAPG problem. Different from existing work applying a backbone network into an entire video frame, AEN involves two parallel pathways in the video visual representation: (i) the agent pathway, which plays at the local level and tells about where agents are and what the agents are doing; (ii) the environment pathway, which plays at the global level and tells about how the environment affects after receiving the actions from the agents as well the relationship between the agents, actions, and the environment. Our experiments demonstrated that AEN outperforms the SOTA methods with C3D backbone on THUMOS-14 and with both C3D and SlowFast backbones on ActivityNet-1.3.

\textbf{Acknowledgment:} 
This material is based upon work supported by the National Science Foundation under Award No. OIA-1946391.
Research is partially supported by Vingroup Innovation Foundation (VINIF) in project code VINIF.2019. DA19, and  the Domestic Master/PhD Scholarship of VINIF.

\textbf{Disclaimer:}
Any opinions, findings, and conclusions or recommendations expressed in this material are those of the author(s) and do not necessarily reflect the views of the National Science Foundation.

%\section{REFERENCES}
\label{sec:refs}

\newpage
\small
\bibliographystyle{IEEEbib}
\bibliography{strings,refs}

\end{document}